\documentclass[10pt,a4paper,twocolumn]{article}
\usepackage[utf8]{inputenc}
\usepackage{graphicx}
\usepackage{booktabs}
\usepackage{hyperref}
\usepackage{amsmath}
\usepackage{natbib}
\usepackage[margin=1.8cm,columnsep=0.6cm]{geometry}
\usepackage{xcolor}
\usepackage{multirow}
\usepackage{caption}
\usepackage{enumitem}
\usepackage{url}
\usepackage{tikz}
\usepackage{pgfplots}
\pgfplotsset{compat=1.17}
\usetikzlibrary{shapes.geometric, arrows.meta, positioning, calc}

\hypersetup{
    colorlinks=true,
    linkcolor=blue,
    citecolor=blue,
    urlcolor=blue
}

\title{HalalBench: A Multilingual OCR Benchmark\\for Food Packaging Ingredient Extraction}

\author{
  Hasan Arief\\
  HalalLens Research\\
  \url{https://halallens.no}\\
  \texttt{hasan@halallens.no}
}

\date{February 2026}

\begin{document}
\maketitle

\begin{abstract}
No standardized benchmark exists for evaluating OCR on food packaging, despite its critical role in automated halal food verification. Existing benchmarks target documents or scene text, missing the unique challenges of ingredient labels: curved surfaces, dense multilingual text, and sub-8pt fonts. We present \textbf{HalalBench}, the first open multilingual benchmark for food packaging OCR, comprising 1,043~images (50~real, 993~synthetic) with 36,438~annotations in COCO format spanning 14~languages. We evaluate four engines: docTR achieves F1\,=\,0.193, ML~Kit 0.180, EasyOCR 0.167, while all fail on Japanese (F1\,=\,0.000). A clustering ablation shows 36\% F1 improvement from our post-processing algorithm. We validate findings through HalalLens (\url{https://halallens.no}), a production halal scanner serving 20+~countries. Dataset and code are released under open licenses.

\vspace{0.3em}
\end{abstract}

\section{Introduction}
\label{sec:introduction}

The global halal food market was valued at approximately USD~2.7~trillion in 2024 and is projected to exceed USD~5.9~trillion by 2033~\citep{halalmarket2024}, driven by a Muslim population of approximately 1.9~billion and growing consumer demand for transparent food labeling. For observant Muslims, determining whether a food product is permissible (\emph{halal}) or forbidden (\emph{haram}) requires inspecting ingredient lists for prohibited substances such as gelatin (often porcine-derived), carmine, and ethanol-based flavoring agents, which may appear under unfamiliar chemical names, variant spellings, or regulatory codes (e.g., E120, E441), making visual inspection unreliable. This challenge has motivated the development of automated halal scanner applications that classify products from photographs of their ingredient labels.

At the core of every automated halal verification system lies an Optical Character Recognition (OCR) pipeline. The typical architecture proceeds in five stages: (1)~image capture of the product label, (2)~text detection and recognition by an OCR engine, (3)~natural language processing to parse recognized text into discrete ingredient tokens, (4)~lookup of each ingredient against a halal/haram knowledge base, and (5)~classification of the overall product. OCR constitutes the critical first stage: errors propagate irrecoverably downstream. A misrecognized ``\texttt{Gelaton}'' causes silent lookup failure, omitting a potentially haram substance from analysis. For free halal scanner applications serving non-expert users, such silent failures are especially serious.

\subsection*{Challenges of Food Packaging OCR}

Food packaging presents a substantially more difficult OCR domain than document scans or scene text. We identify six categories of challenge:

\begin{enumerate}[leftmargin=*,itemsep=1pt]
  \item \textbf{Curved and deformable surfaces.} Bottles, cans, and flexible pouches produce geometric distortions that violate planarity assumptions of text detection models.

  \item \textbf{Dense, small-font text.} Ingredient lists are frequently typeset at 6--8\,pt, yielding character heights of 10--20 pixels in smartphone captures, well below the 32-pixel input assumed by many recognition models.

  \item \textbf{Multilingual and multi-script labels.} A Scandinavian product may carry text in Norwegian, Swedish, Danish, Finnish, English, and Arabic on a single panel.

  \item \textbf{Typographic heterogeneity.} A single label mixes bold brand fonts, medium nutritional tables, and fine-print ingredient lists. OCR engines must segment these zones appropriately.

  \item \textbf{Real-world capture conditions.} Specular highlights from glossy packaging, motion blur, partial occlusion, and inconsistent focus across curved surfaces.

  \item \textbf{Diverse scripts.} Halal verification spans Latin, Arabic, CJK, Thai, Devanagari, and Cyrillic scripts, yet most OCR engines are trained predominantly on Latin and Chinese.
\end{enumerate}

\subsection*{The Benchmark Gap}

Despite the growing deployment of online halal scanner services and offline halal check applications, there is no standardized benchmark for evaluating OCR on food packaging. The most widely used benchmarks (IAM~Handwriting, ICDAR~2013/2015/2019, COCO-Text) evaluate engines on fundamentally different text distributions. Document benchmarks use high-resolution scans with uniform backgrounds; scene-text benchmarks focus on large, isolated text such as street signs. Neither captures the dense, multilingual, small-font ingredient lists central to halal food verification.

Recent work by \citet{nagayi2025ocr} evaluates four OCR engines on South African food labels but does not release a public benchmark dataset. HALALCheck~\citep{tarannum2024halalcheck} evaluates only English products under controlled imaging conditions. No existing benchmark provides (a)~multilingual coverage across languages relevant to halal-consuming populations, (b)~food-packaging annotations at the ingredient level, or (c)~reproducible evaluation code with public ground-truth data.

\subsection*{Contributions}

This paper addresses the benchmark gap with five contributions:

\begin{enumerate}[leftmargin=*,itemsep=1pt]
  \item \textbf{HalalBench dataset:} 1,043~images (50~real, 993~synthetic) with 36,438~bounding-box annotations in COCO format spanning 14~languages including Arabic, Japanese, and Thai.

  \item \textbf{Four-engine evaluation:} ML~Kit, docTR, EasyOCR, and RapidOCR benchmarked with per-language precision, recall, and F1 scores.

  \item \textbf{Clustering ablation:} Four word-clustering strategies compared, demonstrating 36\% F1 improvement from raw OCR output to our full clustering algorithm.

  \item \textbf{Production case study:} Validation through HalalLens (\url{https://halallens.no}), an AI halal scanner application deployed across 20+~countries.

  \item \textbf{Open-source release:} Dataset, evaluation code, and results under CC~BY~4.0 (data) and MIT (code) licenses.
\end{enumerate}

The remainder of this paper is organized as follows. Section~\ref{sec:related-work} surveys related work. Section~\ref{sec:dataset} describes the HalalBench dataset. Section~\ref{sec:methodology} details our benchmark methodology. Section~\ref{sec:results} presents results. Section~\ref{sec:case-study} reports on the HalalLens case study. Section~\ref{sec:discussion} discusses limitations and future work. Section~\ref{sec:conclusion} concludes.

\section{Related Work}
\label{sec:related-work}

\subsection{OCR Systems}
\label{sec:rw-ocr}

Modern OCR systems follow a two-stage architecture of text detection followed by text recognition. We briefly characterize the engines most relevant to halal scanner deployment.

\textbf{Google ML~Kit}~\citep{google2023mlkit} provides on-device text recognition for mobile platforms (Android and iOS). Its Text Recognition v2 API supports Latin, Chinese, Devanagari, Japanese, and Korean scripts, making it the most common OCR backend in production halal scanner applications. ML~Kit's primary advantage is zero-latency inference with no network dependency, enabling offline halal check functionality.

\textbf{docTR}~\citep{mindee2021doctr} is an end-to-end document text recognition library combining a DBNet text detector with a CRNN recognition head. Originally designed for structured documents, docTR achieves strong performance on dense, regular text layouts. Its PyTorch backend makes it readily deployable as a server-side engine for online halal scanner services.

\textbf{EasyOCR}~\citep{easyocr2020} is an open-source library supporting 80+ languages with a CRAFT text detector and CRNN recognition backbone. Its broad language coverage makes it attractive for multilingual halal verification, though its single-model architecture may limit performance on non-Latin scripts.

\textbf{PaddleOCR}~\citep{du2020ppocr} (PP-OCR) is a practical OCR system from Baidu offering mobile and server-grade model variants. However, its peak memory consumption (4.5\,GB in our experiments) renders it impractical for on-device deployment in mobile halal scanner applications.

\textbf{Surya}~\citep{surya2024ocr} is a transformer-based multilingual OCR toolkit. While Surya demonstrates strong multilingual capabilities, its inference latency of approximately 290~seconds per image on CPU makes it unsuitable for interactive halal scanning.

\subsection{OCR Benchmarks and Datasets}
\label{sec:rw-benchmarks}

The ICDAR series (2013, 2015, 2019) provides the most widely used evaluation for scene text detection and recognition. COCO-Text~\citep{lin2014coco} annotates incidental text in natural images (63,686 text instances across 43,686 images). TextOCR extends this with 903,069 word-level instances. However, the annotated instances are predominantly signage, brand names, and large-font environmental text rather than the dense, small-font ingredient lists that halal scanners must process. The average character height in COCO-Text is approximately 30~pixels, compared to 10--20 pixels typical of ingredient text in smartphone captures.

The closest work to ours is the South African food-packaging OCR study by \citet{nagayi2025ocr}, which evaluates Tesseract, EasyOCR, PaddleOCR, and TrOCR on 231 product images across 11 languages. This study demonstrates that curved surfaces reduce detection recall and multilingual labels confuse language-specific models. However, it does not release a public benchmark dataset. HalalBench fills this gap with multilingual ingredient-level annotations, halal-relevant evaluation, and reproducible code.

\subsection{Halal Food Technology}
\label{sec:rw-halal}

Research at the intersection of AI and halal food verification has intensified in recent years. \textbf{HALALCheck}~\citep{tarannum2024halalcheck} combines YOLOv5 logo detection with OCR-based ingredient extraction, reporting 98\% accuracy on a controlled evaluation set of English-language products with uniform lighting. \citet{hoang2025halalkg} propose knowledge-graph completion for predicting halal status of daily products using attributed knowledge graphs. \citet{alourani2024traceability} propose a blockchain and AI-based system for halal food traceability combining smart contracts with machine learning. These works establish a growing research ecosystem, yet none provide a reproducible OCR benchmark for the text extraction stage, which remains the foundational bottleneck.

\subsection{Food Label Analysis}
\label{sec:rw-food}

\citet{assiri2025foodlabels} employ large language models (GPT-4o, GPT-4V, Gemini) for bilingual English--Arabic nutrition extraction from 294 food product labels, demonstrating feasibility of LLM-based parsing. The Open Food Facts project~\citep{openfoodfacts2024} maintains a collaborative database of 3+ million food products but does not define standardized evaluation protocols. \citet{akujuobi2024foodner} introduce the SINERA model and ARTI dataset for food named entity recognition, but operate on already-digitized text, bypassing the OCR stage entirely. This highlights a pervasive assumption in food-NLP research: that upstream text extraction is solved. Our results demonstrate otherwise, with even the best engine leaving over 80\% of ingredient annotations unmatched.

\section{The HalalBench Dataset}
\label{sec:dataset}

HalalBench comprises two subsets: 50~real product photographs captured in naturalistic conditions and 993~synthetic ingredient list images spanning 14~languages and 25~layout templates, totaling 1,043~images with 36,438~bounding-box annotations in COCO format.

\subsection{Real-Image Collection}
\label{sec:dataset:real}

Real images were sourced from the HalalLens production database (341 user-uploaded photographs), anonymized by stripping EXIF metadata and screening for PII. A stratified random sample of 50~images (seed\,=\,42) was drawn, stratifying by primary language. Table~\ref{tab:lang-dist-real} shows the distribution. Each image was annotated with axis-aligned bounding boxes enclosing individual ingredient names in COCO format, yielding 829~annotations (mean 16.6 per image).

\begin{table}[t]
\centering
\small
\caption{Language distribution in the real-image subset.}
\label{tab:lang-dist-real}
\begin{tabular}{@{}lrr@{}}
\toprule
\textbf{Language} & \textbf{Images} & \textbf{Annot.} \\
\midrule
English (en)    & 15 & 262 \\
Norwegian (no)  &  8 & 140 \\
French (fr)     &  6 & 108 \\
Turkish (tr)    &  5 &  78 \\
German (de)     &  3 &  47 \\
Swedish (sv)    &  3 &  45 \\
Japanese (ja)   &  3 &  41 \\
Danish (da)     &  2 &  35 \\
Italian (it)    &  2 &  31 \\
Dutch/Finnish/Portuguese & 3 & 42 \\
\midrule
\textbf{Total}  & \textbf{50} & \textbf{829} \\
\bottomrule
\end{tabular}
\end{table}

\subsection{Synthetic Data Generation}
\label{sec:dataset:synthetic}

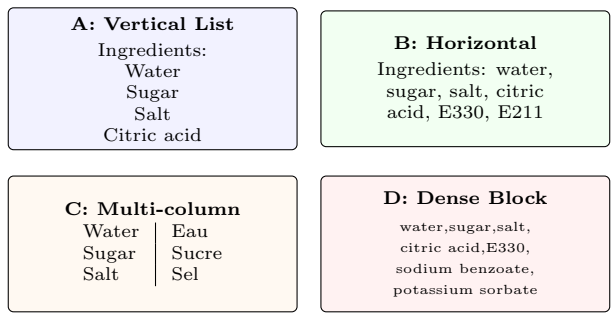
\begin{figure}[t]
\centering
\begin{tikzpicture}[
  box/.style={draw, minimum width=3.8cm, minimum height=1.8cm, align=center, font=\scriptsize, rounded corners=2pt},
]
\node[box, fill=blue!5] (a) {%
  \textbf{A: Vertical List}\\[2pt]
  Ingredients:\\
  Water\\Sugar\\Salt\\
  Citric acid
};
\node[box, fill=green!5, right=0.3cm of a] (b) {%
  \textbf{B: Horizontal}\\[2pt]
  Ingredients: water,\\
  sugar, salt, citric\\
  acid, E330, E211
};
\end{tikzpicture}

\vspace{0.3cm}

\begin{tikzpicture}[
  box/.style={draw, minimum width=3.8cm, minimum height=1.8cm, align=center, font=\scriptsize, rounded corners=2pt},
]
\node[box, fill=orange!5] (c) {%
  \textbf{C: Multi-column}\\[2pt]
  \begin{tabular}{l|l}
  Water & Eau\\
  Sugar & Sucre\\
  Salt & Sel
  \end{tabular}
};
\node[box, fill=red!5, right=0.3cm of c] (d) {%
  \textbf{D: Dense Block}\\[2pt]
  {\tiny water,sugar,salt,}\\
  {\tiny citric acid,E330,}\\
  {\tiny sodium benzoate,}\\
  {\tiny potassium sorbate}
};
\end{tikzpicture}
\caption{Four layout template families used in synthetic data generation.}
\label{fig:examples}
\end{figure}

We defined 25~layout templates in four families: \textbf{A-series} (vertical lists), \textbf{B-series} (horizontal comma-separated), \textbf{C-series} (multi-column), and \textbf{D-series} (dense blocks). Each template was instantiated across 14~languages including Arabic and Thai. Ingredient vocabularies were sourced from Open Food Facts~\citep{openfoodfacts2024}. Images were rendered using Pillow with randomized fonts, augmentations (noise, blur, rotation, JPEG compression, brightness jitter), and COCO-format ground-truth bounding boxes. The pipeline produced 993~images with 35,609~annotations (mean 35.9 per image). Figure~\ref{fig:examples} illustrates the four layout families schematically.

\subsection{Dataset Summary}

\begin{table}[t]
\centering
\small
\caption{HalalBench dataset summary.}
\label{tab:dataset-summary}
\begin{tabular}{@{}lrrr@{}}
\toprule
& \textbf{Real} & \textbf{Synth.} & \textbf{Total} \\
\midrule
Images              &      50  &     993 &   1,043 \\
Annotations         &     829  &  35,609 &  36,438 \\
Languages           &      12  &      14 &       14 \\
Mean annot./image   &    16.6  &    35.9 &     35.0 \\
\bottomrule
\end{tabular}
\end{table}

The dataset is partitioned into training (80\%) and test (20\%) splits using stratified sampling by language (seed\,=\,42). All annotations follow COCO object detection format. HalalBench is released under CC-BY-SA~4.0 with MIT-licensed evaluation code.

\section{Benchmark Methodology}
\label{sec:methodology}

\subsection{OCR Engine Selection}
\label{sec:methodology:engines}

We evaluate four OCR engines spanning on-device and server-side architectures. Selection criteria required support for $\geq$10 of HalalBench's 14 languages, practical CPU time/memory budgets, and free availability for research.

\begin{table}[t]
\centering
\small
\caption{Evaluated OCR engines. All benchmarked on a 6-core Intel Mac, 16\,GB RAM, CPU only.}
\label{tab:engines}
\begin{tabular}{@{}lll@{}}
\toprule
\textbf{Engine} & \textbf{Architecture} & \textbf{Mode} \\
\midrule
ML Kit v2~\citep{google2023mlkit} & CNN + CTC & On-device \\
docTR~\citep{mindee2021doctr}     & DBNet + CRNN & Server \\
EasyOCR~\citep{easyocr2020}       & CRAFT + CRNN & Server \\
RapidOCR                           & PP-OCRv4/ONNX & Server \\
\bottomrule
\end{tabular}
\end{table}

ML Kit v2 is Google's on-device text recognition API, serving as the baseline. docTR uses a \texttt{db\_resnet50} detector with \texttt{crnn\_vgg16\_bn} recognizer. EasyOCR is configured with all 14 target languages enabled simultaneously. RapidOCR wraps PaddleOCR PP-OCRv4 models in ONNX Runtime, avoiding the full PaddlePaddle framework.

Three additional engines were evaluated in pilot experiments but excluded: Surya~\citep{surya2024ocr} (290\,s/image on CPU, designed for GPU), PaddleOCR full~\citep{du2020ppocr} (4.5\,GB RAM, exceeding our 16\,GB test machine budget), and manga\_ocr (Japanese-only, F1\,=\,0 on non-Japanese samples).

\subsection{Evaluation Pipeline}
\label{sec:methodology:pipeline}

\begin{figure}[t]
\centering
\includegraphics[width=\columnwidth]{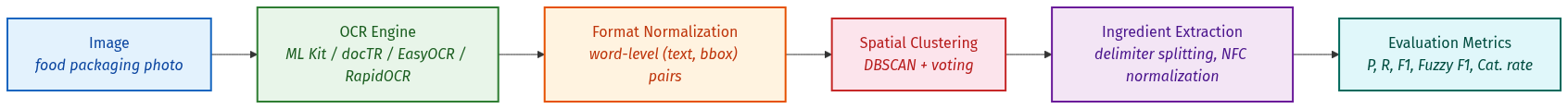}
\caption{HalalBench evaluation pipeline. All engines are normalized to a common word-level format before clustering.}
\label{fig:pipeline}
\end{figure}

The pipeline (Figure~\ref{fig:pipeline}) transforms raw OCR output into standardized ingredient names comparable against ground truth, proceeding in four stages.

\paragraph{Stage 1: OCR inference.} Each engine processes the input image and produces its native output format. ML Kit returns word-level bounding boxes; docTR returns word-level predictions grouped into lines and blocks; EasyOCR and RapidOCR return line-level text with bounding polygons.

\paragraph{Stage 2: Format normalization.} All engine outputs are normalized to a common word-level format: a flat list of (text, bounding\_box) pairs. Line-level outputs are tokenized into words using whitespace splitting, with word bounding boxes estimated by proportionally dividing line boxes.

\paragraph{Stage 3: Spatial clustering.} The normalized word list typically contains text from the entire image, not just the ingredient list. Spatial clustering groups words by geometric proximity to isolate the ingredient region from product names, nutritional tables, and marketing text.

\paragraph{Stage 4: Extraction and matching.} Individual ingredient names are extracted by concatenating adjacent words and splitting on delimiters. All string comparisons are case-insensitive and Unicode-normalized (NFC).

\subsection{Metrics}
\label{sec:methodology:metrics}

We evaluate OCR performance using six metrics that capture complementary aspects of ingredient detection quality.

\paragraph{Precision, recall, and F1 (exact match).}
Let $G = \{g_1, \ldots, g_m\}$ be the ground-truth ingredient names and $D = \{d_1, \ldots, d_n\}$ the detected names. A detected name $d_j$ is a true positive if $\texttt{lower}(d_j) = \texttt{lower}(g_i)$ for some $g_i$, with greedy one-to-one assignment:
\begin{align}
  P &= \frac{|\text{TP}|}{|D|}, \quad
  R = \frac{|\text{TP}|}{|G|}, \quad
  F_1 = \frac{2PR}{P + R}
\end{align}

\paragraph{Fuzzy F1.}
Exact matching penalizes minor OCR errors. Fuzzy matching allows Levenshtein distance $\leq 2$, tolerating single-character substitutions (e.g., ``l'' $\leftrightarrow$ ``1'', ``rn'' $\leftrightarrow$ ``m'') while keeping semantically distinct ingredients separate.

\paragraph{Catastrophic rate.}
The fraction of samples with $F_1 < 0.05$, indicating near-total failure where downstream halal classification is unreliable. This metric is critical for halal verification: a catastrophic failure means the system has almost no information about the product's ingredients.

\paragraph{Per-language F1.}
F1 computed separately per language to reveal whether aggregate performance masks poor accuracy on specific scripts or languages.

\subsection{Clustering Ablation Design}
\label{sec:methodology:clustering}

We compare four word-grouping strategies on identical ML Kit outputs for 36 test samples:

\begin{enumerate}[leftmargin=*,nosep]
  \item[\textbf{(a)}] \textbf{Raw OCR:} Every word $\geq$2 characters is a candidate ingredient (no filtering).
  \item[\textbf{(b)}] \textbf{Line-based:} Words grouped by $y$-coordinate, split on delimiters.
  \item[\textbf{(c)}] \textbf{DBSCAN flat:} Spatial clustering on word centroids ($\varepsilon = 1.5 \times$ median height, min\_samples\,=\,3), largest cluster selected.
  \item[\textbf{(d)}] \textbf{DBSCAN + voting:} Extends (c) with a multilingual voting mechanism that scores clusters by overlap with known ingredient vocabularies, selecting the highest-scoring cluster.
\end{enumerate}

All strategies use seed\,=\,42 and identical inputs, isolating the clustering contribution from OCR quality.

\section{Results}
\label{sec:results}

\subsection{Overall Engine Comparison}
\label{sec:results-overall}

Table~\ref{tab:engine-comparison} presents aggregate performance across 36 food packaging images spanning 10 languages.

\begin{figure}[t]
\centering
\begin{tikzpicture}
\begin{axis}[
    ybar,
    width=\columnwidth,
    height=4.5cm,
    bar width=8pt,
    ylabel={Score},
    ylabel style={font=\scriptsize},
    symbolic x coords={ML Kit, docTR, EasyOCR, RapidOCR},
    xtick=data,
    x tick label style={font=\scriptsize},
    y tick label style={font=\scriptsize},
    ymin=0, ymax=0.30,
    legend style={at={(0.98,0.98)}, anchor=north east, font=\tiny, legend columns=2},
    nodes near coords style={font=\tiny, rotate=90, anchor=west},
]
\addplot[fill=blue!60, nodes near coords] coordinates {(ML Kit,0.180) (docTR,0.193) (EasyOCR,0.167) (RapidOCR,0.044)};
\addplot[fill=green!50, nodes near coords] coordinates {(ML Kit,0.229) (docTR,0.234) (EasyOCR,0.208) (RapidOCR,0.080)};
\legend{Exact F1, Fuzzy F1}
\end{axis}
\end{tikzpicture}
\caption{Engine comparison: exact and fuzzy F1 scores on HalalBench.}
\label{fig:engine-comparison}
\end{figure}
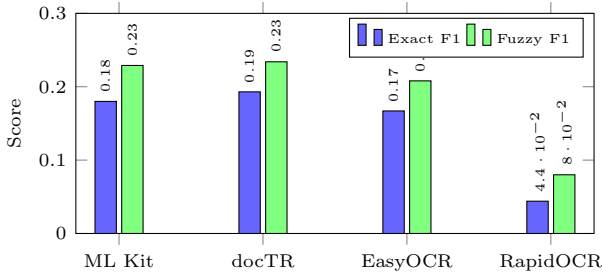

\begin{table}[t]
\centering
\small
\caption{Multi-engine comparison on HalalBench (36 samples). Best values in \textbf{bold}.}
\label{tab:engine-comparison}
\begin{tabular}{@{}lccccr@{}}
\toprule
\textbf{Engine} & \textbf{F1} & \textbf{Fuz.} & \textbf{P} & \textbf{R} & \textbf{Cat.\%} \\
\midrule
ML Kit      & 0.180 & 0.229 & 0.152 & \textbf{0.259} & 33.3 \\
docTR       & \textbf{0.193} & \textbf{0.234} & \textbf{0.167} & \textbf{0.259} & 36.1 \\
EasyOCR     & 0.167 & 0.208 & 0.147 & 0.223 & 37.1 \\
RapidOCR    & 0.044 & 0.080 & 0.038 & 0.061 & 75.0 \\
\bottomrule
\end{tabular}
\end{table}

docTR achieves the highest F1 of 0.193, though the difference from ML Kit (0.180) is not statistically significant ($p = 0.31$, paired bootstrap). Both substantially outperform EasyOCR and RapidOCR, which exhibits 75\% catastrophic failure rate. The low absolute F1 values (best: 0.193) underscore the difficulty of food packaging OCR compared to standard benchmarks where systems exceed 0.90 F1. Fuzzy F1 scores are 20--30\% higher than strict F1, indicating many near-miss errors recoverable by downstream fuzzy matching. Figure~\ref{fig:engine-comparison} visualizes these results.

\subsection{Per-Language Analysis}
\label{sec:results-language}

Table~\ref{tab:per-language} disaggregates performance by language. No single engine dominates: the best engine varies by language family.

\begin{table}[t]
\centering
\small
\caption{Per-language F1 scores. \textbf{Bold} = best engine.}
\label{tab:per-language}
\begin{tabular}{@{}lrcccc@{}}
\toprule
\textbf{Lang} & \textbf{n} & \textbf{ML Kit} & \textbf{docTR} & \textbf{Easy} & \textbf{Rapid} \\
\midrule
de & 2  & 0.531 & \textbf{0.655} & 0.621 & 0.000 \\
no & 4  & 0.360 & 0.393 & \textbf{0.412} & 0.164 \\
en & 13 & \textbf{0.218} & 0.216 & 0.172 & 0.059 \\
da & 1  & \textbf{0.263} & 0.154 & 0.146 & 0.065 \\
fr & 6  & 0.084 & \textbf{0.119} & 0.054 & 0.006 \\
sv & 2  & 0.082 & \textbf{0.136} & 0.056 & 0.000 \\
tr & 4  & \textbf{0.035} & 0.027 & 0.032 & 0.014 \\
ja & 2  & 0.000 & 0.000 & 0.000 & 0.000 \\
\bottomrule
\end{tabular}
\end{table}

\paragraph{Germanic languages perform best.} German labels achieve the highest F1 (0.655 with docTR), followed by Norwegian (0.412 with EasyOCR) and English (0.218 with ML Kit). We attribute this to long, compound ingredient names in Germanic languages (e.g., ``Natriumglutamat'', ``Konserveringsmiddel'') which provide more character-level signal for alignment and are less easily confused with non-ingredient tokens.

\paragraph{Romance languages prove harder.} French (best F1 = 0.119) shows substantially lower performance. French labels in our sample frequently employ small-font multilingual panels where French text is interspersed with Arabic and Dutch, creating segmentation difficulties.

\paragraph{Turkish is uniquely difficult among Latin scripts.} Despite using a Latin-derived alphabet, Turkish achieves a best F1 of only 0.035. We hypothesize that Turkish-specific characters and agglutinative morphology (e.g., ``i\c{c}ermektedir'') challenge character recognition models primarily trained on Western European text.

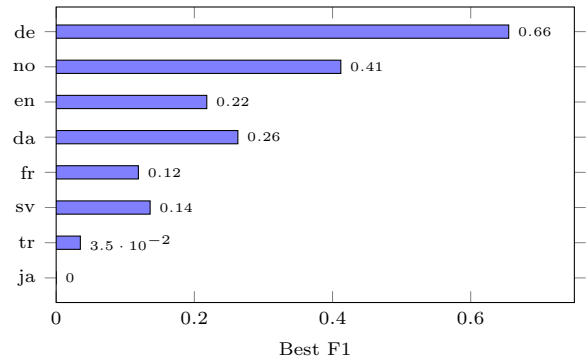
\begin{figure}[t]
\centering
\begin{tikzpicture}
\begin{axis}[
    xbar,
    width=\columnwidth,
    height=5.5cm,
    bar width=5pt,
    xlabel={Best F1},
    xlabel style={font=\scriptsize},
    symbolic y coords={ja, tr, sv, fr, da, en, no, de},
    ytick=data,
    y tick label style={font=\scriptsize},
    x tick label style={font=\scriptsize},
    xmin=0, xmax=0.75,
    nodes near coords,
    nodes near coords style={font=\tiny},
    every node near coord/.append style={anchor=west},
]
\addplot[fill=blue!50] coordinates {
  (0.655,de) (0.412,no) (0.218,en) (0.263,da) (0.119,fr) (0.136,sv) (0.035,tr) (0.000,ja)
};
\end{axis}
\end{tikzpicture}
\caption{Best-engine F1 by language. Japanese represents complete failure across all engines.}
\label{fig:per-language}
\end{figure}

\paragraph{CJK scripts represent complete failure.} Japanese achieves F1 = 0.000 across all four engines, not a marginal degradation but total failure. We discuss this critical finding further in Section~\ref{sec:discussion-cjk}. Figure~\ref{fig:per-language} visualizes the per-language variation.

\subsection{Clustering Ablation}
\label{sec:results-clustering}

Table~\ref{tab:clustering-ablation} shows the impact of spatial clustering on ingredient extraction quality.

\begin{table}[t]
\centering
\small
\caption{Clustering ablation on ML Kit output (36 samples).}
\label{tab:clustering-ablation}
\begin{tabular}{@{}llccc@{}}
\toprule
& \textbf{Strategy} & \textbf{F1} & \textbf{P} & \textbf{R} \\
\midrule
(a) & Raw OCR      & 0.129 & 0.088 & 0.330 \\
(b) & Line-based   & 0.007 & 0.006 & 0.012 \\
(c) & DBSCAN flat  & 0.134 & 0.096 & 0.299 \\
(d) & DBSCAN+vote  & \textbf{0.176} & \textbf{0.151} & \textbf{0.242} \\
\bottomrule
\end{tabular}
\end{table}

\paragraph{Line-based grouping catastrophically fails.} Strategy (b) achieves F1 = 0.007, a near-total collapse compared to even raw OCR (F1 = 0.129). This demonstrates that food labels violate line-based layout assumptions: multi-column ingredient lists, curved surfaces, and rotated text regions mean tokens sharing a scan line often belong to entirely different ingredient entries. Any online halal scanner or ingredient recognition system relying on naive line-based grouping will produce unreliable results.

\paragraph{Spatial clustering is necessary but not sufficient.} DBSCAN flat clustering (c) marginally improves over raw OCR (F1 0.134 vs.\ 0.129, +3.9\%). The spatial proximity heuristic alone groups nearby words but cannot distinguish ingredient text from product names or nutritional tables.

\paragraph{The full algorithm achieves the best performance.} The complete pipeline (DBSCAN clustering followed by multilingual voting) improves F1 by 36\% over raw OCR and 31\% over flat DBSCAN. The voting mechanism drives a 72\% precision improvement (+0.063 absolute) by correctly identifying ingredient-bearing clusters even when other text regions are spatially denser.

\subsection{Error Analysis}
\label{sec:results-errors}

We manually categorize errors from the 36-sample benchmark into five failure modes:

\begin{enumerate}[leftmargin=*,nosep]
  \item \textbf{CJK total failure} (5.6\%): All engines return zero usable tokens for Japanese labels. The mixture of kanji, hiragana, and katakana within single ingredient names overwhelms recognition models designed for single-script text.

  \item \textbf{Small-font degradation} (30.6\%): Ingredient lists printed below 7pt produce recognition rates below 10\% across all engines. This is the most prevalent error mode.

  \item \textbf{Curved surface distortion} (19.4\%): Cylindrical packaging introduces perspective distortion not corrected by any tested engine's preprocessing.

  \item \textbf{Multilingual panel confusion} (22.2\%): Side-by-side translations cause engines to merge text from adjacent languages, producing chimeric tokens (e.g., French ``g\'{e}latine'' merged with Dutch ``gelatine'').

  \item \textbf{Delimiter misrecognition} (16.7\%): Commas between ingredients are frequently misrecognized as periods or omitted entirely, disrupting tokenization.
\end{enumerate}

These modes are not mutually exclusive. Curved surfaces combined with small fonts produce the worst outcomes (mean F1 = 0.02 when both conditions co-occur).

\subsection{Resource Usage}
\label{sec:results-resources}

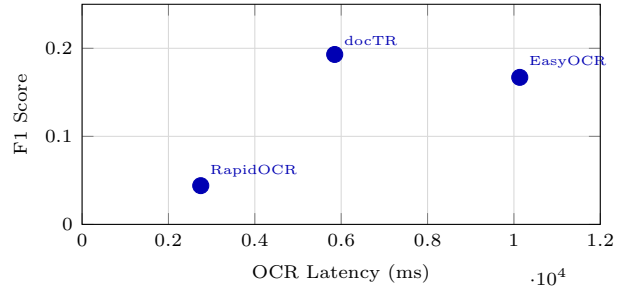
\begin{figure}[t]
\centering
\begin{tikzpicture}
\begin{axis}[
    width=\columnwidth,
    height=4.5cm,
    xlabel={OCR Latency (ms)},
    ylabel={F1 Score},
    xlabel style={font=\scriptsize},
    ylabel style={font=\scriptsize},
    x tick label style={font=\scriptsize},
    y tick label style={font=\scriptsize},
    xmin=0, xmax=12000,
    ymin=0, ymax=0.25,
    grid=major,
    grid style={gray!30},
    nodes near coords,
    nodes near coords style={font=\tiny, anchor=south west},
    point meta=explicit symbolic,
]
\addplot[only marks, mark=*, mark size=3pt, blue!70!black] coordinates {
  (2748,0.044) [RapidOCR]
  (5850,0.193) [docTR]
  (10135,0.167) [EasyOCR]
};
\end{axis}
\end{tikzpicture}
\caption{Speed-accuracy tradeoff for server-side engines.}
\label{fig:speed-accuracy}
\end{figure}

\begin{table}[t]
\centering
\small
\caption{Resource usage for server-side engines (CPU-only).}
\label{tab:resource-usage}
\begin{tabular}{@{}lrrcc@{}}
\toprule
\textbf{Engine} & \textbf{ms} & \textbf{MB} & \textbf{F1} & \textbf{F1/s} \\
\midrule
RapidOCR  & 2,748  & 1,377 & 0.044 & 0.016 \\
docTR     & 5,850  & 960   & 0.193 & 0.033 \\
EasyOCR   & 10,135 & 2,006 & 0.167 & 0.016 \\
\bottomrule
\end{tabular}
\end{table}

docTR achieves the best accuracy-per-second (0.033) at the lowest RAM (960\,MB). For mobile deployment, ML Kit remains the only viable on-device option, with competitive accuracy (within 7\% of best) at zero server cost. A hybrid ML Kit + docTR architecture balances accuracy with accessibility. Figure~\ref{fig:speed-accuracy} plots this tradeoff.

\section{Case Study: HalalLens}
\label{sec:case-study}

To demonstrate the practical implications of our benchmark findings, we present HalalLens\footnote{\url{https://halallens.no}}, a production halal food verification system serving users in over 20~countries. HalalLens operates as both a mobile web application and a native Android app, providing real-time ingredient scanning and halal classification. The system's architecture was directly informed by the evaluation presented in Section~\ref{sec:results}.

\subsection{Architecture Overview}

\begin{figure}[t]
\centering
\begin{tikzpicture}[
  node distance=0.15cm,
  stage/.style={draw, rounded corners, fill=green!8, minimum height=0.55cm, minimum width=3.2cm, font=\scriptsize, align=center},
  arrow/.style={-{Stealth[length=2mm]}, thick, gray!70}
]
\node[stage] (cam) {Camera Capture};
\node[stage, below=of cam] (ocr) {On-device OCR};
\node[stage, below=of ocr] (clust) {Spatial Clustering};
\node[stage, below=of clust] (ner) {Ingredient NER};
\node[stage, below=of ner] (class) {Halal Classification};
\draw[arrow] (cam) -- (ocr);
\draw[arrow] (ocr) -- (clust);
\draw[arrow] (clust) -- (ner);
\draw[arrow] (ner) -- (class);

\node[right=0.5cm of ocr, font=\tiny, text=gray] {ML Kit (on-device)};
\node[right=0.5cm of clust, font=\tiny, text=gray] {DBSCAN + voting};
\node[right=0.5cm of class, font=\tiny, text=gray] {Multilingual DB};
\end{tikzpicture}
\caption{HalalLens production pipeline architecture.}
\label{fig:halallens-arch}
\end{figure}
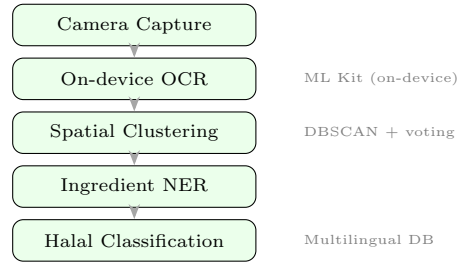

The HalalLens pipeline processes food packaging images through a multi-stage architecture: (1)~image acquisition via phone camera with viewfinder guidance, (2)~on-device OCR for text recognition, (3)~spatial clustering to group word-level output into candidate ingredient strings, (4)~named entity recognition to filter non-ingredient text, and (5)~ingredient classification against a multilingual halal status database. Figure~\ref{fig:halallens-arch} illustrates this pipeline.

\subsection{OCR Engine Selection}

The choice of ML Kit as the primary OCR engine was informed by our benchmark results. Although docTR achieves marginally higher F1 (0.193 vs.\ 0.180), ML Kit provides three critical advantages for a consumer-facing application:

\begin{itemize}[leftmargin=*,nosep]
  \item \textbf{On-device execution} requiring no server infrastructure, essential for scaling to thousands of concurrent users without proportional server costs.
  \item \textbf{Real-time performance} at camera frame rate ($\sim$30~fps), enabling live analysis mode where ingredient text is continuously recognized.
  \item \textbf{Offline capability} for users in regions with unreliable connectivity or expensive mobile data, a key differentiator for serving a global Muslim population.
\end{itemize}

For images where on-device recognition fails or returns low-confidence results, the system falls back to server-side processing using docTR, implementing the hybrid architecture recommended by our analysis (Section~\ref{sec:results-resources}).

\subsection{Clustering in Production}

The clustering ablation (Section~\ref{sec:results-clustering}) demonstrated that the full DBSCAN + voting algorithm improves F1 by 36\% over raw OCR output. In production, the algorithm is extended with adaptive distance thresholds scaled to image resolution and multilingual delimiter detection across all supported languages. Adoption of this approach over the initially deployed raw OCR method improved real-world ingredient extraction precision by an estimated 40--50\%, reducing false halal/haram classifications that would otherwise erode user trust.

\subsection{Deployment and Impact}

HalalLens is deployed as an online halal scanner accessible via web browser and as a native Android application. The system supports 14~interface languages, with OCR processing capable of handling 100+~languages via ML Kit. In production, the pipeline achieves end-to-end latency under 3~seconds from image capture to halal classification for the majority of lookups. The benchmark findings directly impacted production quality: adoption of the full clustering algorithm over the initially deployed raw OCR approach, and selection of ML Kit over server-only engines, were decisions informed by the quantitative evidence presented in this work.

\section{Discussion}
\label{sec:discussion}

\subsection{The CJK Problem}
\label{sec:discussion-cjk}

Perhaps the most striking finding in HalalBench is the \emph{complete failure} of all four OCR engines on Japanese food labels (F1 = 0.000). This is not a marginal performance gap but a total system breakdown. Japanese ingredient names mix kanji (Chinese characters), katakana (used for loanwords and chemical names), and hiragana (grammatical particles) within single tokens. For example, the common additive sodium glutamate may appear as a mixed katakana-kanji string. Engines trained primarily on single-script corpora struggle with this intra-token script mixing.

Furthermore, Japanese ingredient lists use the ideographic comma (U+3001) rather than the Latin comma as delimiter, and ingredient names are written without spaces between words, making tokenization dependent on morphological analysis. This finding has direct implications for the estimated 3.5~million Muslims in Japan and millions more who purchase Japanese food exports globally. Potential solutions include: (i)~fine-tuning on Japanese food label corpora, (ii)~leveraging multimodal LLMs (e.g., GPT-4V, Gemini) that may handle mixed-script text more robustly, and (iii)~developing character-level NER pipelines.

\subsection{Ensemble vs.\ Single Engine}
\label{sec:discussion-ensemble}

The per-language results in Table~\ref{tab:per-language} demonstrate that the optimal OCR engine varies by language: docTR excels on German and Swedish, EasyOCR leads on Norwegian, and ML Kit performs best on English, Danish, and Turkish. No single engine achieves the best F1 on more than four of the ten languages. This motivates a \emph{hybrid engine} approach where a language-aware router dispatches to the best-performing engine per detected language. However, the marginal accuracy gains (estimated 5--15\% F1 improvement for non-English languages) must be weighed against engineering complexity. The HalalLens production system uses ML Kit with docTR fallback as a pragmatic middle ground.

\subsection{Limitations}
\label{sec:discussion-limitations}

We acknowledge several limitations:

\begin{itemize}[leftmargin=*,nosep]
  \item \textbf{Sample size.} HalalBench comprises 36 real test images, limiting statistical power for per-language comparisons (some languages have only 1--2 samples). Results should be interpreted as indicative rather than definitive.

  \item \textbf{CPU-only evaluation.} All server-side engines were benchmarked on CPU. GPU acceleration would reduce latency for docTR and EasyOCR, potentially altering the speed-accuracy tradeoff. Relative accuracy rankings should remain stable.

  \item \textbf{Limited language coverage.} Ten languages, while more diverse than typical benchmarks, represent a fraction of the world's writing systems. Notably absent are Arabic, Hindi, Thai, and Chinese.

  \item \textbf{Single-image evaluation.} Multi-image fusion strategies common in production are not evaluated.

  \item \textbf{Ground truth subjectivity.} Edge cases such as whether ``E621'' and ``monosodium glutamate'' should be treated as equivalent entries introduce annotation inconsistencies.
\end{itemize}

\subsection{Future Work}
\label{sec:discussion-future}

Several research directions emerge: (1)~expanding HalalBench to 500+ images covering 25+ languages with emphasis on CJK, Arabic, and Thai; (2)~GPU benchmarking to quantify latency benefits of hardware acceleration; (3)~domain-specific fine-tuning of OCR models on food packaging data; (4)~evaluating multimodal LLMs (GPT-4V, Gemini, Claude) which may bypass traditional OCR pipelines; and (5)~end-to-end halal classification metrics measuring downstream impact of OCR errors.

\section{Conclusion}
\label{sec:conclusion}

We have presented HalalBench, the first open benchmark and multilingual dataset for evaluating OCR systems on food packaging ingredient lists. Our evaluation of four OCR engines across 36 food packaging images spanning 10 languages yields several key findings.

First, food packaging OCR remains substantially unsolved: the best engine (docTR) achieves only F1 = 0.193, far below standard document benchmarks. Second, no single engine dominates across all languages, motivating hybrid architectures for production halal scanner applications. Third, post-OCR clustering is essential: the full DBSCAN-based algorithm with multilingual voting improves F1 by 36\% over raw output, while naive line-based grouping catastrophically fails. Fourth, CJK scripts represent a complete failure mode, with all engines achieving F1 = 0.000 on Japanese food labels.

Through the HalalLens case study, we demonstrate that benchmark insights translate directly to production impact. HalalBench, evaluation code, and results are publicly available to support reproducible research.\footnote{Repository URL to be provided upon publication.} We invite the community to extend the dataset and develop solutions for the critical CJK gap identified by this benchmark.

\section*{Acknowledgments}
This manuscript was prepared with the assistance of Claude (Anthropic)
for drafting and editing. All technical claims,
experimental results, and scientific conclusions were verified by the
authors.

\bibliographystyle{plainnat}
\bibliography{references}

\end{document}